\documentclass[preprint,12pt]{elsarticle}

\usepackage{amssymb}
\usepackage{amsmath}
\usepackage{algorithm}
\usepackage{multirow} % Required for multirowsib}

\usepackage{url}
\usepackage{verbatim}
\usepackage{booktabs}  % 支持 \toprule 等命令
\usepackage[table,HTML]{xcolor}
\usepackage{threeparttable}
\usepackage[T1]{fontenc} % 改善字体编码支持
\usepackage{textcomp}
\usepackage{makecell}                 % 三线表-竖线
\usepackage{multicol} 
\usepackage{caption} 
\usepackage{graphicx}
\usepackage{rotating}  % 支持sidewaystable
\usepackage{amssymb}
\usepackage{amsmath}
\usepackage{wasysym}

\usepackage{xcolor} % 可选，用于更多颜色控制
\usepackage{hyperref} % 关键设置
\hypersetup{
hidelinks,
linkcolor=[RGB]{0, 154, 205},
colorlinks=true,
citecolor=[RGB]{0, 154, 205}
}

\journal{Pattern Recognition}

\begin{document}

\begin{frontmatter}



\title{OTSNet: A Unified Observation-Thinking-Spelling Network for Scene Text Recognition}


\author[1]{Lixu Sun}
\ead{sunLixu@stu.xju.edu.cn}
\author[1]{Nurmemet Yolwas \corref{cor1}}
\ead{nurmemet@xju.edu.cn}

\author[1]{Wushour Silamu}
\ead{wushour@xju.edu.cn}

\affiliation[1]{organization={School of Computer Science and Technology},
            addressline={Xinjiang University}, 
            city={Urumqi},
%          citysep={}, % Uncomment if no comma needed between city and postcode
            postcode={830046}, 
            %state={},
            country={China}}
\cortext[cor1]{Corresponding author at: School of Computer Science and Technology of Xinjiang University, Urumqi 830046, China.}

%% Abstract
\begin{abstract}
Scene Text Recognition (STR) remains challenging due to real-world complexities, where decoupled visual-linguistic optimization in existing frameworks amplifies error propagation through cross-modal misalignment. 
Visual encoders exhibit attention bias toward background distractors, while decoders suffer spatial misalignment in parsing geometrically deformed text, collectively degrading recognition accuracy for irregular patterns.
Inspired by the hierarchical cognitive processes in human visual perception, 
we propose a novel three-stage network named OTSNet that embodies a neurocognitive-inspired Observation-Thinking-Spelling pipeline for unified STR modeling.   
The model comprises three core components:
(1) A Dual Attention Macaron Encoder (DAME) that refines visual features through differential attention maps to suppress irrelevant regions and enhance discriminative focus;
(2) Position-Aware Module (PAM) and Semantic Quantizer (SQ), which integrate spatial context with glyph-level semantic abstraction via adaptive sampling;
And (3) A Multi-Modal Collaborative Verifier (MMCV) that enforces self-correction through cross-modal fusion among visual features, semantic features and character features.
Extensive experiments demonstrate that OTSNet achieves state-of-the-art performance, attaining 83.5\% average accuracy on the challenging Union14M-L benchmark and 79.1\% on the heavily occluded OST dataset, establishing new records across 9 out of 14 evaluation scenarios.  

\end{abstract}

\begin{graphicalabstract}
\begin{center} % 替代\centering
\includegraphics[width=0.7\textwidth,keepaspectratio]{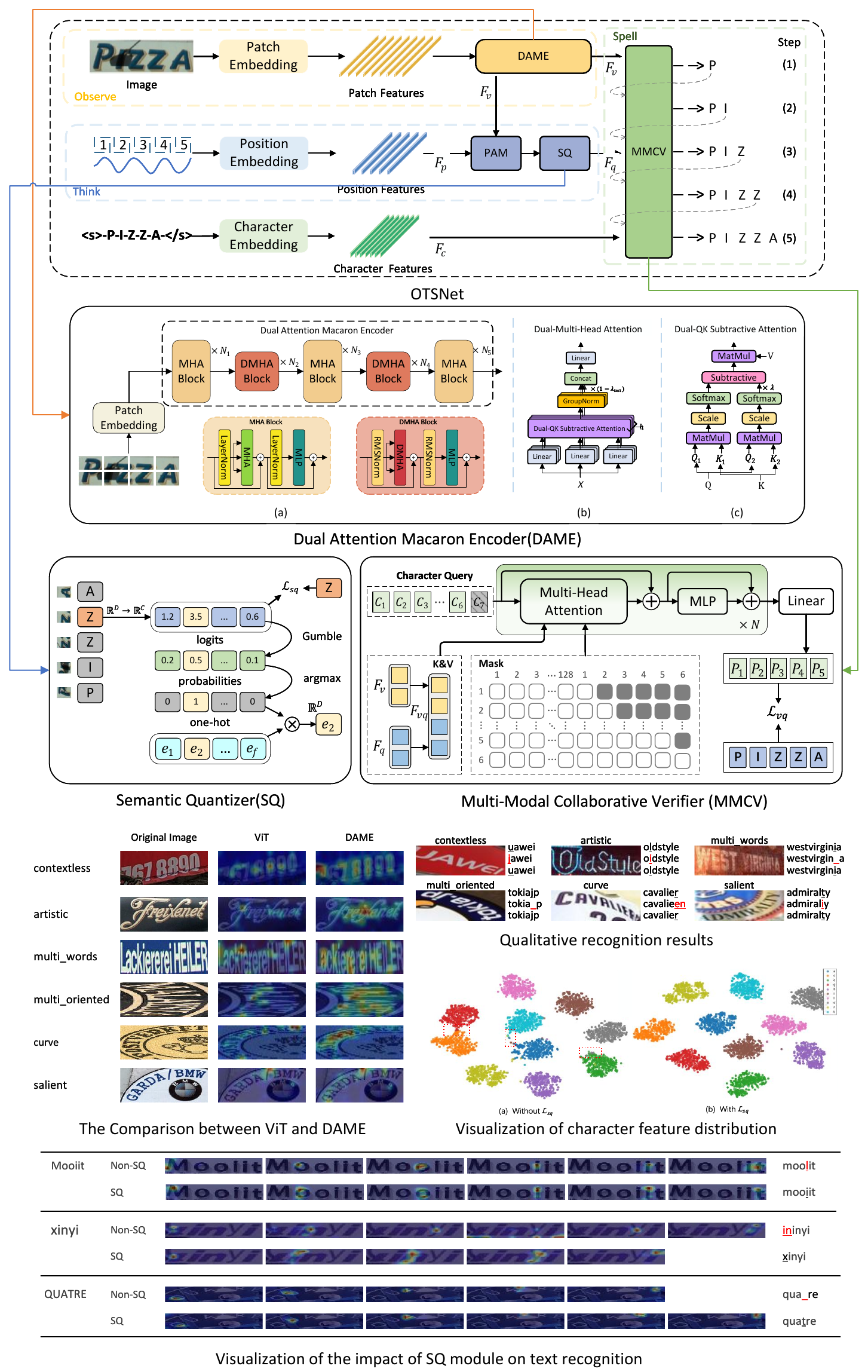}
\par % 确保标题在正确位置
\captionof{figure}{Graphical abstract of OTSNet: A Unified Observation-Thinking-Spelling Network for Scene Text Recognition}
\label{fig:graphical-abstract}
\end{center}
\end{graphicalabstract}

\begin{highlights}
\item Proposed OTSNet: a neurocognitive-inspired pipeline for scene text recognition with unified modeling.
\item DAME relieves encoder attention bias through differential attention maps in visual feature refinement.
\item Integrates Position-Semantic Module boosting robustness to geometric deformation.
\item Novel Multi-modal Verifier (MMCV) enables self-correction via cross-modal fusion.
\item OTSNet achieves SOTA performance on benchmark scene text recognition tasks.
\end{highlights}

%% Keywords

\end{frontmatter}



\section{Introduction}\label{sec1}

Scene Text Recognition (STR), which identifies text from natural scene images,  has garnered significant attention in the field of computer vision.
Diverse textual instances in real-world scenarios, including road signs, posters, billboards, and license plates, 
present substantial challenges due to factors such as curved or perspective text layouts, complex background interference, and diverse typographical variations \cite{1,3}.
Despite extensive research efforts to develop advanced visual feature extractors and character sequence generators, 
existing methods still exhibit limitations in handling realistic challenges including occlusion, blurriness, deformation, and other environmental disturbances.

Early STR methods \cite{6,8} primarily focused on segmenting and classifying individual characters, which neglected semantic information within the text. 
Recent studies \cite{9} have begun leveraging language model (LM) to capture textual semantics, achieving notable progress.
For instance, \cite{9} employed a pre-trained bidirectional LM and decoupled the visual model (VM) from the LM by blocking gradient flow to enhance the iterative optimization phase. 
However, such a decoupled learning strategy may yield erroneous results due to the oversight of visual features, 
and deviates from the coordinated and unified cognitive processes inherent in human text recognition.

To better integrate semantic and visual cues, \cite{VisionLAN} proposed the Vision-Language Attention Network (VisionLAN), 
which employs a language-aware visual mask to occlude targeted character regions during training, thereby strengthening visual feature learning.  
Following this, \cite{wei2024busnet} introduced the Balanced Unified Synchronous Network (BUSNet), 
which treats images as linguistic modalities and mitigates over-reliance on language models by harmonizing visual-linguistic information and learning unified external-internal representations. 
Inspired by these advancements, we raise a novel question: Could there exist a more human-intuitive STR modeling paradigm? 
The fact that humans lacking formal instruction in the target language retain the ability to efficiently transcribe text from scene images, indicating the potential to design more effective STR models through emulation of human cognitive processes.

We posit that the human text recognition process can be decomposed into three stages: observation, thinking, and spelling.
During the initial observation phase, the human visual system rapidly localizes textual regions through holistic perception rather than pixel-wise scanning \cite{ungerleider2000mechanisms}. 
This contrasts with conventional Convolutional Neural Networks (CNNs) \cite{li2024volter} that passively receive information through progressive receptive field expansion,
whereas the patch embedding and self-attention mechanisms in Vision Transformer (ViT) \cite{vits} show active scanning properties that emulate the human capability of holistic structural acquisition during rapid saccades.
However, traditional ViT suffer from attention dispersion-excessive focus on irrelevant regions which degrades discriminative feature representation \cite{zhong2024ndorder}.

In the thinking stage,  humans dynamically adjust attentional focus through saccadic eye movements to perform enhanced sampling of key glyph regions \cite{zheng2024cdistnet}. 
This active perception mechanism effectively handles complex scenarios involving blur and occlusion. 
Furthermore, unsupervised second-language learners rely on grapho-morphological analogical reasoning rather than strict character matching during cognitive processing.

During the spelling phase, the human neural system exhibits a distinctive dynamic self-correction mechanism \cite{yang2024class}. 
This mechanism establishes a collaborative verification system through three aspects: 
(1) Geometry-based reading sequence reconstruction: analyzing spatial attributes such as text arrangement orientation and line spacing distribution to establish a reading sequence;
(2) Semantic-constrained character association modeling: 
constructing contextual prediction models using memory traces formed by preceding contextual information; 
(3) Multimodal feature verification: cross-checking abstract glyph features with original visual inputs to achieve bidirectional verification between character representations and visual information.
This enables iterative refinement through a cyclic hypothesis–feedback loop.

Based on the above analysis, we propose an Observe-Think-Spell cognitive paradigm-based scene text recognition network (OTSNet). 
First, we introduce the Dual Attention Macaron Encoder (DAME), which draws inspiration from the differential operator concept in cybernetics \cite{laplante2018comprehensive,ye2024differential}. 
It optimizes the ViT attention mechanism through attention map differential denoising to generate more accurate visual features. 
Second, the Position-Aware Module (PAM) simulates the cognitive mechanism of dynamic attention focus adjustment in the human visual system \cite{zhao2024decoder}.
Concurrently, the Semantic Quantizer (SQ) captures discrete abstract character units to extract glyph semantic features.
Finally, the Multi-Modal Collaborative Verifier (MMCV) dynamically integrates visual information with glyph semantic features,
introducing character information to produce final predictions. 
Notably, the SQ module enables non-blocking training that preserves model integrity while enhancing performance. 
Additionally, OTSnet achieves competitive results without pretraining.

The contributions of this paper are as follows:

\begin{enumerate}
  \item We propose OTSNet, a cognition-inspired network unifying observation (focus-enhanced visual sampling), thinking (abstract character units discretizing), spelling (multi-modal reasoning). 
  The key innovations include: 
  \begin{itemize}
    \item  The Position-Aware Module (PAM) and Semantic Quantizer (SQ) dynamically project visual features into discrete semantic spaces to improve robustness against ambiguous text.
    \item Multi-Modal Collaborative Verifier (MMCV) enforces cross-modal consistency constraints to reduce over-reliance on single-modal information.
  \end{itemize}
  \item Inspired by cybernetics, we design a Dual Attention Macaron Encoder (DAME) that optimizes ViT via attention map differential denoising, resolving attention deviation. 
  \item Experiments on benchmark datasets show OTSNet's SOTA performance, validating its robustness in complex scene text recognition tasks. 
\end{enumerate}

\section{Related Work}
\subsection{Scene Text Recognition}
With the emergence of deep learning, STR methods can generally be categorized into two classes: segmentation-driven approaches and sequence-based frameworks. 
The former category \cite{zhang2016multi} generally adopts a two-stage pipeline: precise character localization/segmentation from complex backgrounds, succeeded by independent recognition of isolated characters.
This approach requires high-quality character-level annotated data, which often incurs significant annotation costs. 
In contrast, sequence-based approaches formulate STR as an image-to-text sequence transformation problem, which can be further divided into Connectionist Temporal Classification (CTC)-based methods (\cite{ctc}) and attention-based models. 
In CTC frameworks \cite{shi2016end}, 
a canonical implementation employs CNNs for hierarchical visual encoding, RNNs for contextual sequence modeling, and CTC-based sequence-label alignment for decoding invariance.

Recent advances highlight the dominance of attention mechanisms in STR owing to their dynamic feature alignment capabilities \cite{bai2018edit,bautista2022scene}.
For instance, ASTER \cite{shi2018aster} pioneered the integration of Bahdanau attention with bidirectional LSTM decoders, enabling fine-grained character-wise alignment.
Inspired by the success of ViT, ViTSTR \cite{atienza2021vision}  explores the viability of pure-transformer architectures for STR. 
To further enhance model performance, researchers have explored improved integration of linguistic knowledge through solutions like PIMNet \cite{qiao2021pimnet}, SRN \cite{12}, and ABINet \cite{9}. 
PIMNet proposes progressive prediction and similarity distance concepts between non-autoregressive and autoregressive models to learn linguistic knowledge from autoregressive counterparts. 
SRN and ABINet enhance VM outputs through linguistic modalities, with final predictions generated via fusion between VM and LM outputs.

Despite these advancements, existing solutions like SVTR \cite{du2022svtr} and PARSeq \cite{bautista2022scene} exhibit performance limitations due to their asymmetric focus on either visual features or linguistic constraints.
While hybrid models such as ABINet and MATRN \cite{MATRN} attempt to balance modalities, their modular designs lack coherent cross-modal interaction mechanisms. 
To address this gap, VOLTER \cite{li2024volter} proposes a contrastive learning framework with dual-stream architecture to enforce synergy between visual and linguistic representations. 
Concurrently, BUSNet develops an iterative refinement pipeline for progressive optimization of vision-language joint inference. 
However, both frameworks require sufficient pre-training to equip the model with essential prior knowledge for achieving optimal performance.

\subsection{Visual Feature Enhancement Methods}
Recent research advancements have increasingly focused on enhancing visual feature representation to advance STR performance.
These methodologies can be systematically classified into two categories: 
(1) image preprocessing-based approaches that refine input quality to strengthen visual features, 
and (2) architecture-driven strategies that directly augment feature representations to address irregular text challenges.

\subsubsection{Image Preprocessing Methods}
To counteract the detrimental effects of suboptimal imaging conditions, preprocessing techniques have shown significant efficacy. 
\cite{luo2021separating} leveraged GANs to suppress background clutter while preserving textual semantics, thereby boosting recognition robustness. 
Extending super-resolution principles \cite{yang2016consistent,yang2018drfn}, \cite{wang2019textsr} proposed a unified framework integrating super-resolution and recognition modules to recover fine-grained textural details from blurred inputs. 
\cite{li2021character} introduced a geometry-aware rectification mechanism that corrects character-level skewness and rotation via positional-orientation constraints.
\cite{wu2022two} further enhanced curved text recognition through dual-domain correction, jointly optimizing geometric alignment and pixel-wise reconstruction.

\subsubsection{Architectural Innovation Methods} 
To directly model irregular text patterns, specialized network architectures have been extensively explored. 
\cite{lee2020recognizing} designed an adaptive 2D self-attention encoder with local dependency modeling, enabling spatial context capture for arbitrarily shaped text.
\cite{cheng2018aon} proposed a multi-directional feature fusion paradigm, addressing training instabilities in Spatial Transformer Network (STN)-based methods through directional feature aggregation and weighted sequence generation.

Diverging from prior work, we reformulate ViT by integrating differential cybernetics principles. Specifically, we propose a hybrid architecture alternating standard Multi-Head Self-Attention (MHSA) and novel Differential Multi-Head Attention (DMHA) blocks, achieving precise visual feature extraction tailored for STR tasks.

\section{Methodology}
\begin{figure*}[!t]
  \centering
  \includegraphics[width=0.95\textwidth]{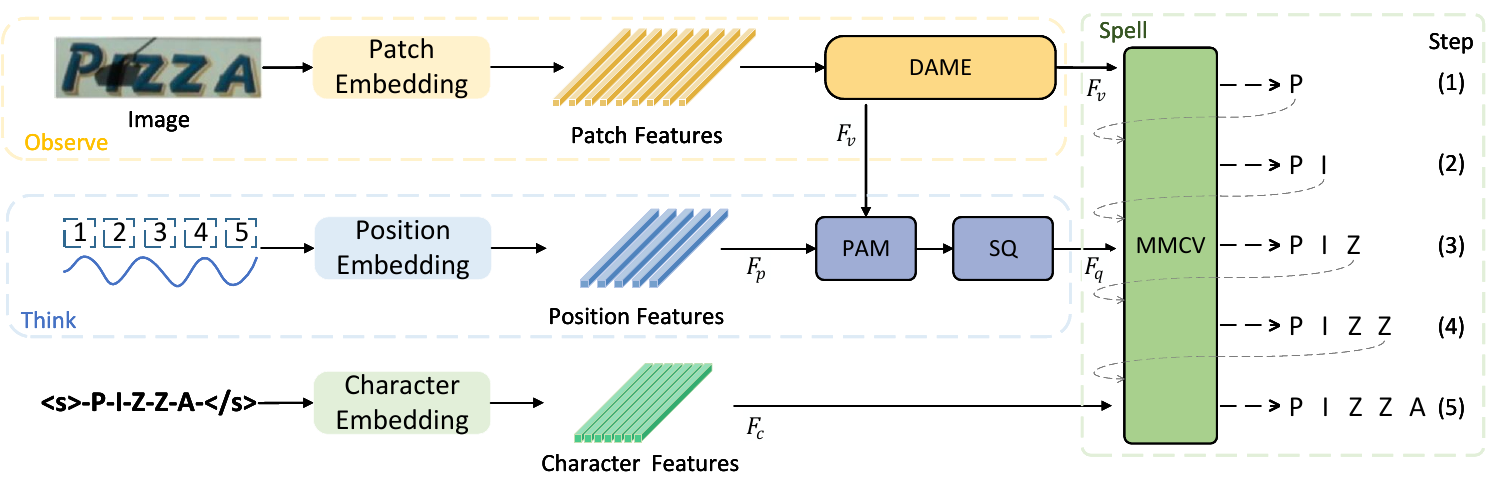}
  \caption{An overview of OTSNet. OTSNet divides the input image into patches and extracts low-level features, followed by the DAME's deep visual feature extraction to capture fine-grained details. The PAM then fuses positional and visual features via SQ to form  glyph semantic features.  Finally, MMCV integrates visual, glyph semantic, and character features for joint modeling, producing the final recognition output.}
  \label{OTSNet}
\end{figure*}

The proposed OTSNet comprises three sequential stages: observation, thinking, and spelling, with its architecture detailed in Figure \ref{OTSNet}.
During the Observation stage, OTSNet extracts image patches from input data, followed by visual feature extraction via the DAME.
In the Thinking stage, OTSNet generates positional information through coordinate encoding.
Then PAM integrates visual features with positional encodings, producing enhanced visual focus features that amplify responses in critical glyph regions for precise character localization.
The subsequent SQ discretizes visual focus features into abstract character units, which are further embedded into glyph semantic features through a learnable codebook.
Finally, in the spelling stage, MMCV integrates visual features, glyph semantic features, and character features through three branches to generate the final prediction results.

\subsection{Observation}
Although the Transformer architecture has achieved breakthrough progress in computer vision due to its global semantic modeling capability and the scalability of pre-training strategies, 
its core self-attention mechanism still faces the attention dispersion phenomenon in complex scenarios.
Specifically, the global computation pattern leads to undesired attention dispersion towards task-irrelevant regions, 
which limits fine-grained feature capture capacity and consequently degrades discriminative accuracy for critical features.

\begin{figure*}[!t]
  \centering
  \includegraphics[width=0.95\textwidth]{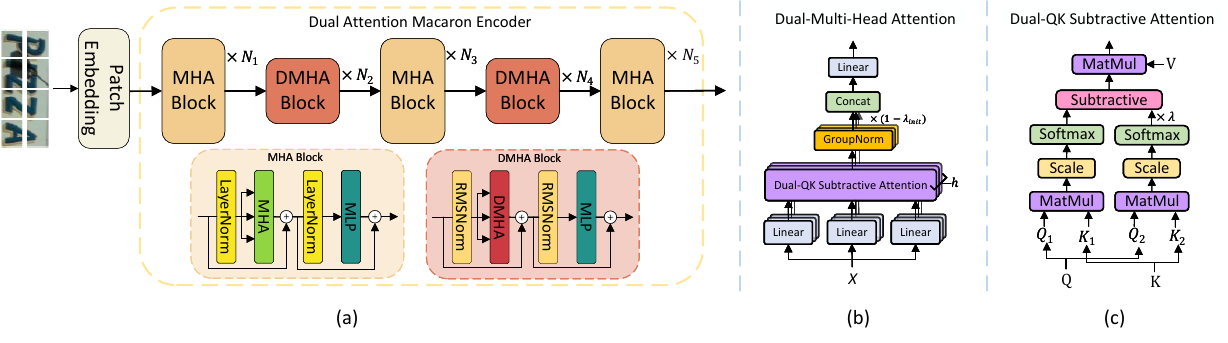}
  \caption{Architecture of the observation stage in OTSNet. (a) The Dual Attention Macaron Encoder (DAME), which interleaves standard MHA and proposed DMHA blocks in a Macaron-style structure. (b) Internal design of the Differential Multi-Head Attention (DMHA) block. (c) The Dual-QK Subtractive Attention mechanism, which enhances local discriminability via subtraction of two independent attention maps.}
  \label{OB}
\end{figure*}

To address this, we propose the Dual Attention Macaron Encoder (DAME), whose architecture is illustrated in Figure~\ref{OB}(a). 
Inspired by the Macaron-Net design~\cite{Conformer}, DAME interleaves standard MHSA blocks and our proposed Differential Multi-Head Attention (DMHA) blocks in a deep hierarchical structure. 
This hybrid arrangement enables a progressive optimization process that alternates between \textit{global contextual modeling} (via MHSA) and \textit{local discriminative refinement} (via DMHA), thereby mitigating the over-suppression issue observed when DMHA is used in isolation.

The MHSA module inherits the residual structure of standard Transformers, with its computational process formalized as:
\begin{align}
&Q  = \mathrm{LayerNorm}(X)W_q, \\
&K  = \mathrm{LayerNorm}(X)W_k, \\
&V  = \mathrm{LayerNorm}(X)W_v, \\
&A  = \mathrm{Softmax}(QK^{\top}/\sqrt{d}), \\
&X'  = X + AV, \\
&X  = X' + \mathrm{MLP}(\mathrm{LayerNorm}(X)),
\end{align}
where $ X \in \mathbb{R}^{B \times N\times D}$ is the input of the block, $W_q$, $W_k$, $W_v$ are the learnable projection matrices, and $d$ denotes the head dimension.

The internal architecture of DMHA is illustrated in Figure~\ref{OB}(b).
While retaining the classical multi-head mechanism,
we introduce a novel Dual-QK Subtractive Attention mechanism to enhance inter-feature discriminability while suppressing redundant information.
Given input features $X \in \mathbb{R}^{B \times N\times D}$, we first generate dual independent projections:
\begin{align}
  &Q_1 = XW_{q1}, \quad Q_2 = XW_{q2}, \\
  &K_1 = XW_{k_1}, \quad K_2 = XW_{k_2}, \quad V = XW_v,
\end{align}
where $Q_1, Q_2, K_1, K_2 \in \mathbb{R}^{N\times D/2}$ and $V \in \mathbb{R}^{N\times D}$. 
Two attention maps are computed as:
\begin{align}
  &A_1 = \mathrm{Softmax}(Q_1K_1^{\top}/\sqrt{d}), \\
  &A_2 = \mathrm{Softmax}(Q_2K_2^{\top}/\sqrt{d}).
\end{align}
The final output is obtained via  subtraction:
\begin{align}
  \mathrm{Output} = (A_1 - \lambda \cdot A_2)V,
\end{align}
where 
\begin{align}
  \lambda = \exp(\lambda_{q_1}^\top \lambda_{k_1}) - \exp(\lambda_{q_2}^\top \lambda_{k_2}) + \lambda_{\mathrm{init}},
\end{align}
and $\lambda_{q_1}, \lambda_{k_1}, \lambda_{q_2}, \lambda_{k_2} \in \mathbb{R}^D$ are learnable vectors.

We adopt the subtraction operation because it draws inspiration from the differential mechanism in cybernetics—by computing the difference between two attention responses, it effectively suppresses redundant activations in noisy regions while amplifying the contrast between text and background, thereby yielding sharper and more discriminative attention distributions.

Given $h$ attention heads, each head employs independent projection matrices $W_{Q}^i$, $W_{K}^i$, $W_V^i$ for $i\in[1,h]$. 
After normalization, the outputs are fused as:
\begin{align}
&\mathrm{head}_i  = \mathrm{DualQK}(X,W_{Q}^i,W_{K}^i,W_V^i,\lambda_{\mathrm{init}}), \\
&\mathrm{head}_i'  = (1 - \lambda_{\mathrm{init}}) \cdot \mathrm{RMSNorm}(\mathrm{head}_i), \\
&\mathrm{MultiHead}(X) = \mathrm{Concat}(\mathrm{head}_1',\cdots ,\mathrm{head}_h')  W_{\mathrm{proj}},
\end{align}
where $W_{\mathrm{proj}}\in \mathbb{R}^{D\times D}$ is a learnable projection matrix, and $h = D/2d$.

We adopt RMSNorm (Root Mean Square Normalization) instead of the conventional LayerNorm in the DMHA block, motivated by the intrinsic mechanism of differential attention. The core of DMHA lies in the subtraction $A_1 - \lambda A_2$, which enhances text-background contrast by amplifying discriminative signals while suppressing background noise. This operation critically relies on the absolute magnitude of feature responses. LayerNorm, by centering features to zero mean, may inadvertently erase such magnitude cues essential for differential computation. In contrast, RMSNorm preserves the global scale of activations by normalizing only with respect to the root mean square, thereby maintaining the intensity information that drives effective contrast enhancement. Moreover, RMSNorm exhibits lower statistical variance and improved training stability compared to LayerNorm, especially when processing local or sparse visual tokens—a common scenario in scene text recognition. Its computational simplicity and compatibility with residual connections further align with the Macaron-style architecture of DAME, facilitating stable gradient flow and efficient optimization.

\subsection{Thinking}
The visual features $F_v \in \mathbb{R}^{B \times N \times D}$ generated by the observation stage are inherently unordered, as they correspond to image patches without explicit alignment to character positions in the output sequence. However, the subsequent Semantic Quantizer (SQ) requires a one-to-one correspondence between its input tokens and the $T$ character slots (positions $1$ to $T$, with $T = 25$ being the maximum sequence length). Without such alignment, SQ would operate on an unstructured set of visual tokens, rendering glyph-level semantic abstraction ill-posed.

To address this, we first generate positional encoding features $F_p \in \mathbb{R}^{B \times T \times D}$ using sinusoidal embeddings~\cite{vaswani2017attention}, which encode absolute positional priors for each character slot.

We then introduce a Position-Aware Module (PAM) that leverages Multi-Head Cross-Attention (MHCA) to align the unordered visual features with the ordered character positions:
\begin{align}
F_u = \mathrm{MHCA}(F_p, F_v, F_v),
\end{align}
where $F_p$ serves as the query, and $F_v$ provides the key and value. This mechanism adaptively aggregates relevant visual regions for each character position, yielding position-aligned visual focus features $F_u \in \mathbb{R}^{B \times T \times D}$.  
Unlike the self-attention used in the observation stage—where attention dispersion may occur due to the lack of positional structure—the queries in MHCA are already structured by position. 
Consequently, the subtraction-based denoising strategy employed in DMHA is both unnecessary with this cross-attention paradigm.

\begin{figure}
  \centering
  \includegraphics[width=2.5in]{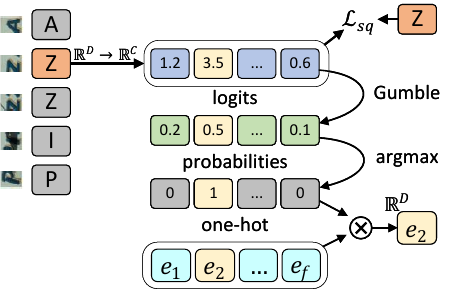}
  \caption{Schematic diagram of the Semantic Quantizer (SQ) workflow.}
  \label{sq}
\end{figure}

Next, the Semantic Quantizer (SQ) (see Figure~\ref{sq}) maps these continuous focus features into discrete, abstract character units to enable semantic abstraction. Specifically, a learnable linear projection $\Phi: \mathbb{R}^D \to \mathbb{R}^C$ compresses and reorganizes the feature space:
\begin{align}
Q = \Phi(F_u; \theta_\Phi),
\end{align}
where $Q \in \mathbb{R}^{B \times T \times C}$ represents logits over $C$ predefined semantic units ($C < D$). This step preserves glyph-level discriminability while decoupling visual appearance from semantic representation.

To enable end-to-end training despite the non-differentiability of hard discretization, we adopt the Gumbel-Softmax relaxation~\cite{jang2016categorical}. For each position $t$, we perturb the logits with i.i.d.\ Gumbel noise $\{G_i\}_{i=1}^C$, where $G_i \sim -\log(-\log U(0,1))$:
\begin{align}
Q'_t = [q_1 + G_1, q_2 + G_2, \dots, q_C + G_C].
\end{align}
A softened categorical distribution is then computed using a temperature parameter $\tau$:
\begin{align}
p_i^{(t)} = \frac{\exp(q'_i / \tau)}{\sum_{j=1}^C \exp(q'_j / \tau)}.
\end{align}
During early training (with large $\tau$), gradients flow smoothly across categories; as $\tau \to 0$, the distribution converges to a one-hot vector:
\begin{align}
\lim_{\tau \to 0} p_i^{(t)} = \mathbb{I}\big[i = \arg\max_j (q_j + G_j)\big].
\end{align}

Finally, the quantized glyph semantic features $F_q \in \mathbb{R}^{B \times T \times D}$ are reconstructed via a learnable codebook $\mathbf{E} \in \mathbb{R}^{C \times D}$:
\begin{align}
F_q = \sum_{i=1}^C p_i^{(t)} \mathbf{E}_i.
\end{align}
This quantization process endows the model with discrete, interpretable semantic units, which serve as the foundation for cross-modal verification in the subsequent spelling stage.

\subsection{Spelling}
\begin{figure}
  \centering
  \includegraphics[width=0.75\linewidth]{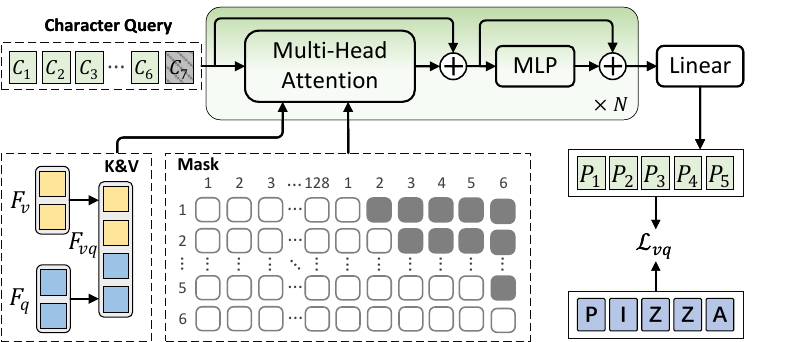}
  \caption{Architecture of the MMCV.}
  \label{mmcv}
\end{figure}

To address cross-modal alignment in open-set scenarios, we propose the Multi-Modal Collaborative Verifier (MMCV). As shown in Figure~\ref{mmcv}, MMCV constructs a joint representation $F_{vq} = \mathrm{Concat}(F_v, F_q) \in \mathbb{R}^{B \times (N + T) \times D}$ by fusing raw visual features $F_v$ and glyph semantic features $F_q$.

The core of MMCV is a triple-interaction attention mechanism:
\begin{align}
\mathrm{CA}(Q, K, V) = \mathrm{Softmax}\left(\frac{QK^\top}{\sqrt{d_k}} \odot M^{\mathrm{attn}}\right) V,
\end{align}
where the query $Q = F_c$ originates from character embeddings, and key-value pairs $(K, V) = F_{vq}$ come from the multi-modal fusion. The causality-preserving mask $M^{\mathrm{attn}}$ enforces autoregressive decoding:
\begin{align}
M^{\mathrm{attn}}_{ij} = 
\begin{cases}
0, & j \leq N + i, \\
-\infty, & j > N + i,
\end{cases}
\end{align}
ensuring that the prediction of the $i$-th character only attends to visual patches ($1\!\sim\!N$) and previously generated semantic units ($1\!\sim\!i$).

Notably, we retain standard self-attention (not DMHA) in MMCV because its role is to model semantic dependencies among characters and cross-modal alignment, not to suppress visual noise. The visual denoising has already been handled in the observation stage, and the glyph semantics from SQ provide clean, discrete inputs. Thus, DMHA is neither necessary nor suitable here.

During training, character features $F_c \in \mathbb{R}^{B \times L \times D}$ are derived from ground-truth labels via Word2Vec-based embeddings ($L = T = 25$). During inference, $F_c$ is built autoregressively from previously predicted characters, enabling open-vocabulary recognition without pre-specifying text length.

\subsection{Optimization objectives}

OTSNet is optimized in an end-to-end manner by integrating visual features, 
glyph features, and character features through a multi-task cross-entropy objective. 
To explicitly bridge the learning objectives of positional awareness and semantic quantization, 
we introduce the semantic quantization loss ${L}_{sq}$, 
which enables the model to better focus on the relationship abstract character units and visual focal features.
The overall objective function is formulated as:
\begin{equation}
  {L} = {L}_{vq} + \alpha_1 {L}_{sq}
\end{equation}
where $\alpha_1$ is a hyper-parameter controlling the trade-off, empirically set to 0.3. 
This formulation establishes a balanced optimization framework that coordinates feature representations across different semantic levels.

\section{Experiments}
\subsection{Dataset and Experimental Details}
We evaluate OTSnet on multiple benchmarks spanning diverse scenarios:
\begin{enumerate}
  \item{Conventional benchmarks (Common benchmarks) : ICDAR 2013 (IC13) (\cite{karatzas2013icdar}), Street View Text (SVT) (\cite{7}), IIIT5K Word (IIIT5K) (\cite{IIIT5K}), ICDAR 2015 (IC15) (\cite{karatzas2015icdar}), Street View Text Perspective (SVTP) (\cite{phan2013recognizing}), and CUTE80( \cite{risnumawan2014robust}). We use the versions containing 857 and 1,811 images for IC13 and IC15 respectively.}
  \item{The test set of the Union14M-L benchmarks (Union14M benchmarks) (\cite{jiang2023revisiting}): A challenging benchmark containing seven subsets—Curved (Cur), Multi-Oriented (M-O), Artistic Font (Art), Contextless (Con), Salient (Sal), Multi-Word (M-W), and General (Gen).}
  \item{Occluded Scene Text (OST) (\cite{VisionLAN}): Divided into Mild Occlusion (OSTw) and Heavy Occlusion (OSTh) subsets based on occlusion severity.}
\end{enumerate}
Owing to its higher sample difficulty, broader scene coverage, and more systematically organized categorization, the Union14M-L benchmark provides a more rigorous evaluation framework that is particularly conducive to advancing robust scene text recognition research. 
Therefore, this study places particular emphasis on accuracy performance within this benchmark.

\begin{sidewaystable}[!p]
  \caption{Accuracy comparison of OTSNet with other methods on different benchmarks}
  \centering
  \footnotesize
  \definecolor{FFC7}{HTML}{FFFFC7}  % 定义浅黄色
  \definecolor{DAEEFF}{HTML}{DAEEFF} 
  \definecolor{Gray}{HTML}{D3D3D3}
  \setlength{\tabcolsep}{2.5pt}
  \begin{tabular}{ccccccccccccccc}
    \toprule
    \cellcolor{FFC7} IIIT5k  & {\cellcolor{FFC7}SVT} & \cellcolor{FFC7}{IC13} & \cellcolor{FFC7} IC15 & \cellcolor{FFC7} SVTP & \cellcolor{FFC7} CUTE80& || & 
    \cellcolor{DAEEFF}Curve &\cellcolor{DAEEFF} Multi-Oriented & \cellcolor{DAEEFF}Artistic & \cellcolor{DAEEFF}Contextless & \cellcolor{DAEEFF}Salient & \cellcolor{DAEEFF}Multi-Words & \cellcolor{DAEEFF}General \\
  \end{tabular}
  \begin{tabular}{c |c |*{6}{c} c |*{7}{c} c| c |c |c }
    % \toprule
    \toprule
    % 第一层表头（跨两行）
    Method & 
    Venue & 
    \multicolumn{6}{c}{\cellcolor{FFC7} {Common Benchmarks}} & 
    Avg & 
    \multicolumn{7}{c}{\cellcolor{DAEEFF} {Union14M Benchmarks}} & 
    Avg & 
    OST& 
    Size& 
    FPS \\
    \midrule
    CRNN (\cite{shi2016end})     & TPAMI16 & 82.9 & 81.6 & 91.1 & 69.4 & 70.0 & 65.5 & 76.75 & 48.1 & 13.0 & 51.2 & 62.3 & 41.4 & 60.4 & 68.2 & 49.24 & 58.0 & \textbf{16.20} & 172 \\
    ASTER  (\cite{shi2018aster})   & TPAMI19 & 96.1 & 93.0 & 94.9 & 86.1 & 87.9 & 92.0 & 91.68 & 70.9 & 82.2 & 56.7 & 62.9 & 73.9 & 58.5 & 76.3 & 68.75 & 61.9 & 19.04 & 67.1 \\
    NRTR (\cite{sheng2019nrtr})     & ICDAR19 & 98.1 & 96.8 & 97.8 & 88.9 & 93.3 & 94.4 & 94.89 & 67.9 & 42.4 & 66.5 & 73.6 & 66.4 & 77.2 & 78.3 & 67.46 & 74.8 & 44.26 & 17.3 \\
    DAN (\cite{wang2020decoupled})      & AAAI20  & 97.5 & 94.7 & 96.5 & 87.1 & 89.1 & 94.4 & 93.24 & 74.9 & 63.3 & 63.4 & 70.6 & 70.2 & 71.1 & 76.8 & 70.05 & 61.8 & 27.71 & 99.0 \\
    SRN  (\cite{yu2020towards})    & CVPR20  & 97.2 & 96.3 & 97.5 & 87.9 & 90.9 & 96.9 & 94.45 & 78.1 & 63.2 & 66.3 & 65.3 & 71.4 & 58.3 & 76.5 & 68.43 & 64.6 & 51.70 & 67.1 \\
    SEED  (\cite{11})    & CVPR20  & 96.5 & 93.2 & 94.2 & 87.5 & 88.7 & 93.4 & 92.24 & 69.1 & 80.9 & 56.9 & 63.9 & 73.4 & 61.3 & 76.5 & 68.87 & 62.6 & 23.95 & 65.4 \\
    RoScanner (\cite{yue2020robustscanner}) & ECCV20  & 98.5 & 95.8 & 97.7 & 88.2 & 90.1 & 97.6 & 94.65 & 79.4 & 68.1 & 70.5 & 79.6 & 71.6 & 82.5 & 80.8 & 76.08 & 68.6 & 47.98 & 64.1 \\
    ABINet$^*$  (\cite{9})  & CVPR21  & 98.5 & 98.1 & 97.7 & 90.1 & 94.1 & 96.5 & 95.83 & 80.4 & 69.0 & 71.7 & 74.7 & 77.6 & 76.8 & 79.8 & 75.72 & 75.0 & 36.86 & 73.0 \\
    VisionLAN (\cite{VisionLAN}) & ICCV21  & 98.2 & 95.8 & 97.1 & 88.6 & 91.2 & 96.2 & 94.50 & 79.6 & 71.4 & 67.9 & 73.7 & 76.1 & 73.9 & 79.1 & 74.53 & 66.4 & 32.88 & 93.5 \\
    PARSeq$^*$  (\cite{bautista2022scene})  & ECCV22  & 98.1 & 96.3 & 97.9 & 89.2 & 92.1 & 96.2 & 95.00 & 83.5 & 83.2 & 70.4 & 78.3 & 80.7 & 78.9 & 83.1 & 79.73 & 78.1 & 23.83 & 19 \\
    MATRN$^*$ ( \cite{MATRN})   & ECCV22  & \textbf{98.8} & \textbf{98.3} & 97.9 & 90.3 & 95.2 & 97.2 & \textbf{96.29} & 82.2 & 73.0 & 73.4 & 76.9 & 79.4 & 77.4 & 81.0 & 77.62 & 77.8 & 44.34 & 46.9 \\
    MGP-STR (\cite{MGP-STR})  & ECCV22  & 97.9 & 97.8 & 97.1 & 89.6 & 95.2 & 96.9 & 95.74 & 85.2 & 83.7 & 72.6 & 75.1 & 79.8 & 71.1 & 83.1 & 78.65 & 78.7 & 148.00 & 120 \\
    SVTR  (\cite{du2022svtr})    & IJCAI22 & 98.0 & 97.1 & 97.3 & 88.6 & 90.7 & 95.8 & 94.58 & 76.2 & 44.5 & 67.8 & 78.7 & 75.2 & 77.9 & 77.8 & 71.17 & 69.6 & 18.09 & 161 \\
    LPV-B  (\cite{LPV-B})   & IJCAI23 & 98.6 & 97.8 & 98.1 & 89.8 & 93.6 & \textbf{97.6} & 95.93 & 86.2 & 78.7 & 75.8 & 80.2 & 82.9 & 81.6 & 82.9 & 81.20 & 77.7 & 30.54 & 82.6 \\
    LISTER  (\cite{LISTER})  & ICCV23  & \textbf{98.8} & 97.5 & \textbf{98.6} & 90.0 & 94.4 & 96.9 & 95.48 & 78.7 & 68.8 & 73.7 & \textbf{81.6} & 74.8 & 82.4 & 83.5 & 77.64 & 77.1 & 51.11 & 44.6 \\
    CDistNet (\cite{zheng2024cdistnet}) & IJCV24  & 98.7 & 97.1 & 97.8 & 89.6 & 93.5 & 96.9 & 95.59 & 81.7 & 77.1 & 72.6 & 78.2 & 79.9 & 79.7 & 81.1 & 78.62 & 71.8 & 43.32 & 15.9 \\
    BUSNet$^*$  (\cite{wei2024busnet})   & AAAI24  & 98.3 & 98.1 & 97.8 & \textbf{90.2} & \textbf{95.3} & 96.5 & 96.06 & 83.0 & 82.3 & 70.8 & 77.9 & 78.8 & 71.2 & 82.6 & 78.10 & 78.7 & 32.10 & 83.3 \\
    OTE  \cite{OTE}     & CVPR24  & 98.6 & 96.6 & 98.0 & 90.1 & 94.0 & 97.2 & 95.74 & 86.0 & 75.8 & 74.6 & 74.7 & 81.0 & 65.3 & 82.3 & 77.09 & 77.8 & 20.28 & 55.2 \\
    \rowcolor{Gray} OTSNet    & -       & 98.1 & 96.6 & 98.4 & \textbf{90.2} & 94.4 & 97.2 & 95.82 & \textbf{87.2} & \textbf{87.7} & \textbf{76.7} & \textbf{81.6} & \textbf{83.6} & \textbf{82.9} & \textbf{84.8} & \textbf{83.50} & \textbf{79.1} & 28.6 & 79.2 \\
    \bottomrule
  \end{tabular}
  \begin{tablenotes}   
    \footnotesize
    \item All the models are trained on U14M-Filter. Size denotes the model size$(M)$. FPS is uniformly measured on one NVIDIA 1080Ti GPU. $^*$ indicates the reproduced results. Bold values denote the first accuracy in each column.
  \end{tablenotes}
  \label{tab:sota}
\end{sidewaystable}

The training data is sourced from the real-world Union14M dataset. 
To address data leakage caused by overlapping samples between the U14M training subset and test subset, we adopt the filtered version U14M-Filter (\cite{du2024svtrv2}), which excludes overlapping data.

Training is performed using the AdamW optimizer with a weight decay of 0.05, a batch size of 512, and an initial learning rate of $5\times 10^{-4}$. 
We employ a single-cycle learning rate scheduler with 1.5 epochs of linear warmup over 20 total epochs. 
Data augmentation techniques including rotation, perspective distortion, motion blur, and Gaussian noise are randomly applied, with a maximum text length of 25. 
In the DAME architecture, we configure a 12-layer Macron structure with $N_1=2$, $N_2=1$, $N_3=6$, $N_4=1$ and $N_5=2$, initialized with $\lambda_{init}=0.05$. 
For MMCV,  we set $N=3$.
The character vocabulary contains 96 classes, and all models are trained on two NVIDIA A40 GPUs using mixed-precision computation.

\subsection{Comparison with State-of-the-arts}

To show the effectiveness of OTSNet, we conducted comparative experiments with 18 popular scene text recognition methods.
All models were trained on the U14M-Filter dataset, with results presented in Table~\ref{tab:sota}. 
Across 14 evaluation scenarios, OTSNet achieved the best performance in 9 scenarios.
Particularly, OTSNet establishes a new state-of-the-art accuracy of 83.5\% on the seven challenging subsets of the Union14M-L dataset.
On the Curved (Cur) and Multi-Oriented (M-O) datasets, OTSNet significantly outperforms prior methods such as SVTR~\cite{du2022svtr} and CDistNet~\cite{zheng2024cdistnet}, which implicitly model geometric deformation through local attention or feature aggregation but lack explicit structural reasoning.
By integrating joint visual-semantic character modeling, OTSNet achieves 79.1\% accuracy on the OST dataset, ranking first among existing methods. 

Although OTSNet does not attain top performance on every subset of the Common Benchmarks (e.g., slightly behind ABINet on IIIT5K), its average accuracy of 95.82\% remains highly competitive. 
This slight gap on clean, short-text benchmarks is a deliberate trade-off: OTSNet prioritizes robustness to visual ambiguity, occlusion, and distortion in complex scenes, whereas models like ABINet leverage strong linguistic priors that excel when visual cues are clear. 

\begin{figure}
  \centering
  \includegraphics[width=0.8\linewidth]{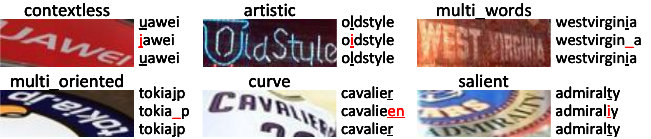}
  \caption{Qualitative recognition results that BUSNet fails but OTSNet successes. For each image,the right-side annotation comprises three components: (1) Ground-truth text (top), (2) BUSNet predictions with red-colored error highlighting (middle), and (3) Our OTSNet predictions (bottom). }
  \label{otsvsbusnet}
\end{figure}

Compared with BUSNet, OTSNet demonstrates absolute accuracy gains of 4.2\% (Cur), 5.4\% (M-O), 4.9\% (Art), 3.7\% (Con), 4.8\% (Sal), 11.7\% (M-W), and 2.2\% (Gen), respectively. 
Qualitative analysis in Figure~\ref{otsvsbusnet} further illustrates OTSNet’s superior robustness in recognizing challenging samples with distorted fonts or complex backgrounds, where BUSNet fails.

Collectively, these results validate that OTSNet’s cognition-inspired Observation–Thinking–Spelling paradigm achieves a more balanced and robust integration of visual perception and semantic reasoning, setting a new standard for scene text recognition in real-world, visually complex environments.

\subsection{Ablation Study}
We conduct a series of ablation studies to comprehensively analyze the performance of the proposed OTSNet. 
First, we validate the effectiveness of the DAME module and evaluate the impact of different $\lambda_{init}$  values. 
Second, we investigate the contributions of the PAM, SQ, and MMCV. 
Finally, we assessed the influence of ${L}_{sq}$ on the overall model performance. 
All experiments are conducted on the same training set to ensure fairness and comparability.

\subsubsection{The Effect of DAME}
To examine the efficacy of the DAME module, we conduct ablation studies as shown in Table~\ref{tab:DAME}.
The baseline ViT model achieves an average accuracy of 82.89\% on the Union14M benchmarks.
When all MHA blocks are replaced with DMHA (i.e., using DMHA in isolation), performance drops significantly to 77.83\%.
This degradation confirms that the subtraction-based attention mechanism in DMHA, while effective at suppressing background noise, tends to over-suppress informative visual cues when applied without global contextual guidance.

\begin{figure}
  \centering
  \includegraphics[width=0.8\linewidth]{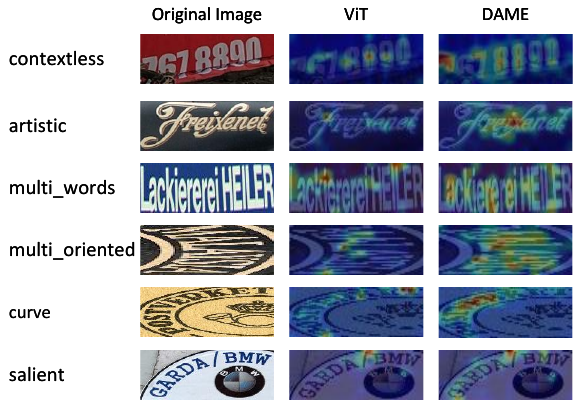}
  \caption{Comparison of attention maps on Union14M images between ViT and DAME. Left: original image; middle: ViT attention; right: DAME attention. DAME exhibits sharper focus on glyph regions and reduced background distraction.}
  \label{vitvsdiff}
\end{figure}

In contrast, our full DAME—constructed by interleaving standard MHA and DMHA blocks within a Macaron-style architecture—achieves 83.50\% accuracy, outperforming both the ViT baseline and the DMHA-only variant.
This improvement stems from a balanced optimization process: MHA preserves holistic semantic dependencies across the entire image, while DMHA refines local discriminability through differential attention.
Their alternation within the Macaron framework prevents feature collapse and enables effective global-local feature interaction.
As illustrated in Figure~\ref{vitvsdiff}, DAME produces attention maps that are more precisely focused on text regions and less distracted by complex backgrounds compared to the baseline ViT.

\begin{table}[!t]
  \caption{Ablation Study on the Effectiveness of DAME}
  \centering
  % \small
  % \setlength{\tabcolsep}{5pt}
  \begin{tabular}{cccccccccc}
    \toprule
    D & M  & Cur & M-O & Art & Con & Salt & M-W & Gen & Avg \\
    \midrule
    - & -  & 86.2 & 87.0 & 76.1 & 81.0 & 83.6 & 81.6 & 84.7 & 82.89 \\
    $\checkmark$ & -  & 81.0 & 81.4 & 69.1 & 75.6 & 78.9 & 76.0 & 82.8 & 77.83 \\
    $\checkmark$ & $\checkmark$ & 87.2 & 87.7 & 76.7 & 81.6 & 83.6 & 82.9 & 84.8 & \textbf{83.50} \\
    \bottomrule
  \end{tabular}
  \begin{tablenotes}
    \footnotesize
    \item D: use DMHA; M: use Macaron architecture (interleaved MHA-DMHA stacking); $\checkmark$: enabled.
  \end{tablenotes}
  \label{tab:DAME}
\end{table}

Furthermore, we evaluate the sensitivity to the initialization value $\lambda_{\mathrm{init}}$, as shown in Table~\ref{tab:lamda}.
Performance remains stable across different settings, with the best result achieved at $\lambda_{\mathrm{init}} = 0.05$.
We attribute this robustness to the learnable scaling parameters $\lambda_{q_1}, \lambda_{k_1}, \lambda_{q_2}, \lambda_{k_2}$, which dynamically adjust the effective $\lambda$ during training, allowing the model to adapt to diverse text patterns and background complexities.

\begin{table}[!t]
  \caption{Ablation Study on the Impact of $\lambda_{\mathrm{init}}$}
  \centering
  % \small
  % \setlength{\tabcolsep}{5pt}
  \begin{tabular}{ccccccccc}
    \toprule
    $\lambda_{\mathrm{init}}$ & Cur & M-O & Art & Con & Salt & M-W & Gen & Avg \\
    \midrule
    0.05 & 87.2 & 87.7 & 76.7 & 81.6 & 83.6 & 82.9 & 84.8 & \textbf{83.50} \\
    0.10 & 87.6 & 87.0 & 76.2 & 82.8 & 84.3 & 81.3 & 84.8 & 83.43 \\
    0.15 & 88.1 & 88.0 & 75.3 & 81.4 & 84.0 & 80.4 & 84.7 & 83.13 \\
    \bottomrule
  \end{tabular}
  \label{tab:lamda}
\end{table}

\subsubsection{The Effect of PAM, SQ and MMCV}
To evaluate the effectiveness of the residual modules PAM, SQ, and MMCV, we conducted ablation experiments as shown in Table~\ref{tab:PAM}.
When only PAM is enabled (without SQ or MMCV), the model achieves 77.91\% average accuracy, with notably low performance on Art (69.2\%) and M-W (73.4\%), 
indicating that positional alignment alone is insufficient for robust recognition.
 
\begin{table}[!t]
  \caption{Ablation Study About The Effectiveness Of  PAM, SQ and MMCV}
  \centering
  % \small
  % \setlength{\tabcolsep}{2pt}
  \begin{tabular}{ccccccccccc}
    \toprule
    PAM&MMCV& SQ& Cur & M-O & Art & Con & Salt & M-W & Gen & Avg \\
    \midrule
    $\checkmark$& - & - & 82.8 & 80.5 & 69.2 & 78.3 & 79.3 & 73.4 & 81.9 & 77.91 \\
    - & $\checkmark$& - &  87 & 87 & 75.3 & 81.5 & 83.6 & 82 & 84.6 & 83.00 \\
    $\checkmark$& $\checkmark$& - &  86 & 85.8 & 73.2 & 81.4 & 83.9 & 79.5 & 84.3 & 82.01 \\
    $\checkmark$& $\checkmark$& $\checkmark$&  87.2 & 87.7 & 76.7 & 81.6 & 83.6 & 82.9 & 84.8 & \textbf{83.50} \\
    \bottomrule
    \end{tabular}
    \begin{tablenotes}   
      \footnotesize
      \item $\checkmark$means using.
    \end{tablenotes}
    \label{tab:PAM}
\end{table}

After introducing the MMCV module, the average accuracy in improved by 5.1 percentage points, with 6.1-point and 8.6-point gains specifically observed on the Art and M-W subsets, respectively. 
This improvement primarily stems from MMCV's incorporation of semantic features that not only focus on character positions but also encode partial text sequence information, 
thereby enhancing the model's capability to handle complex cases.

Notably, when combining PAM with MMCV, we observed an unexpected 1\% decline in overall accuracy. 
We attribute this to functional redundancy between the modules, as both prioritize positional feature extraction, leading to suboptimal multimodal feature fusion.
Finally, incorporating the SQ module established a progressive pipeline ("spatial localization → character semantic abstraction → multimodal fusion"), achieving an average accuracy of 83.5\% in challenging scenarios. 

The Figure \ref{sq_important} below illustrates the inference flow with/without SQ, where heatmap regions show prediction rationale during character recognition. 
As shown, OTSNet with SQ accurately locates character regions even in complex images, enabling reliable predictions. 
Experimental results confirm that the SQ not only alleviates text sequence conflicts through glyph semantic guidance but also complements MMCV's character sequence features via its glyph modeling capability, 
jointly addressing challenging STR problems.

\begin{figure*}
  \centering
  \includegraphics[width=0.95\textwidth]{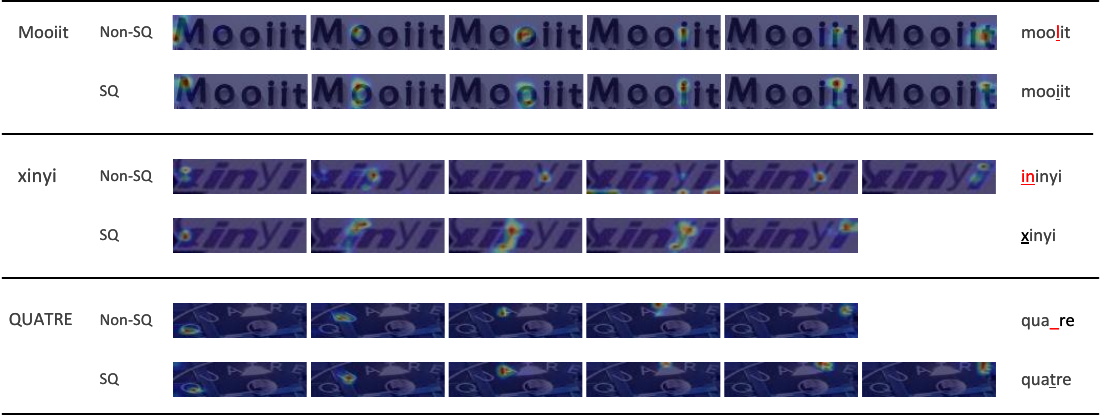}
  \caption{Visualization of the impact of SQ module on text recognition.}
  \label{sq_important}
\end{figure*}

\subsubsection{The Components of SQ}
To evaluate SQ, we compare the experimental results of multiple components in SQ as presented in the table \ref{tab:others}.  
The introduction of  $L_{sq}$ enhances the model's overall performance by an average of 1.03\%, with particularly notable improvements on the Art dataset, indicating its effectiveness complex texture features. 
By integrating Gumbel-Softmax—which employs a continuous relaxation-based gradient estimation mechanism—with the guidance of  $L_{sq}$, the model achieves optimal average performance of 83.5\%.
Furthermore, the results reveal distinct sensitivity patterns across datasets: The Art dataset exhibits greater reliance on $L_{sq}$'s  detail reconstruction capability, while the M-W dataset benefits from Gumbel-Softmax's robust discrete modeling properties. 
The combination of Gumbel-Softmax and $L_{sq}$ achieves optimal performance by optimizing the synergy between discretization and loss function design, thereby validating its effectiveness in multi-task scenarios.
\begin{table}[!t]
  \caption{Ablation Study About The Components Of SQ}
  \centering
  % \small
  % \setlength{\tabcolsep}{3pt}
  \begin{tabular}{cccccccccc}
    \toprule
    Method & $L_{sq}$ & Cur & M-O & Art & Con & Salt & M-W & Gen & Avg \\
    \midrule
    - & - & 86   & 85.8 & 73.2 & 81.4 & 83.9 & 79.5 & 84.3 & 82.01 \\
    Normal         & - & 86.2 & 87.2 & 74.8 & 81.8 & 83.1 & 80.3 & 84.5 & 82.56 \\
    - & $\checkmark$ & 87.0 & 87.5 & 77.4 & 81.9 & 83.2 & 79.6 & 84.7 & 83.04 \\
    Normal         & $\checkmark$ & 85.8 & 87.7 & 75.8 & 81.1 & 82.8 & 81.3 & 84.5 & 82.70 \\
    Detach         & $\checkmark$ & 86.6 & 86.9 & 75.9 & 81.8 & 82.6 & 81.8 & 84.7 & 82.90\\
    GS & $\checkmark$ & 87.2 & 87.7 & 76.7 & 81.6 & 83.6 & 82.9 & 84.8 & \textbf{83.50} \\
    \bottomrule
    \end{tabular}
    \begin{tablenotes}   
      \footnotesize
      \item Normal: basic mapping; Detach: detach final layer output before codebook mapping; GS: uses Gumbel-Softmax in codebook mapping; $\checkmark$ indicates the method is used.
    \end{tablenotes}
    \label{tab:others}
\end{table}

In the table \ref{tab:lq}, we investigate the impact of $L_{sq}$. 
The consistent peak performance at $\alpha_1 = 0.3$ across diverse subsets (Cur, M-O, Art, M-W, etc.) suggests that this weighting strikes an optimal balance between visual fidelity and semantic abstraction. 
A smaller $\alpha_1$ (e.g., 0.2) provides insufficient gradient signal to align the quantized semantic units with visual focal features, leading to suboptimal codebook utilization. 
Conversely, a larger $\alpha_1$ (e.g., 0.4) overemphasizes semantic consistency at the expense of visual detail preservation, causing the model to ignore subtle glyph variations—particularly detrimental on the M-W subset where inter-word spacing and character deformation require fine-grained visual cues.

More importantly, the semantic quantization loss $L_{\mathrm{sq}}$ actively shapes the learning dynamics of the SQ by enforcing a semantic--visual alignment constraint. 
Specifically, $L_{\mathrm{sq}}$ is defined as the cross-entropy between the Gumbel-Softmax logits $Q$ and the ground-truth character labels, encouraging each semantic unit in the learnable codebook $\mathbf{E}$ to specialize in a distinct glyph pattern.
 This transforms the SQ from a generic feature compressor into a discriminative glyph encoder, 
 thereby enabling the downstream MMCV to perform reliable cross-modal verification even under severe occlusion or distortion.

\begin{figure}
  \centering
  \includegraphics[width=0.75\linewidth]{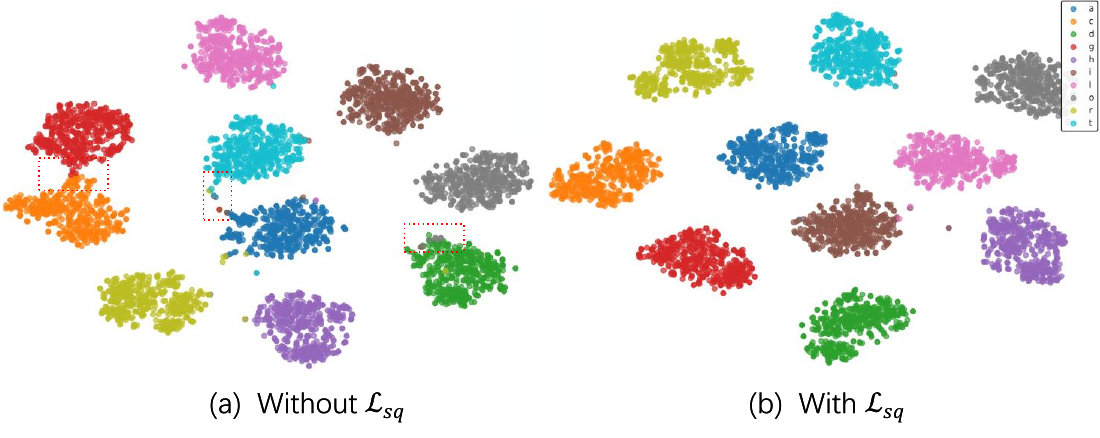}
  \caption{Visualization of character feature distribution.}
  \label{lsq}
\end{figure}

To validate this effect, we employ t-SNE to project the glyph-level features into 2D space, 
focusing on 10 frequently confused character categories. As shown in Figure~\ref{lsq}(a), without $L_{\mathrm{sq}}$, 
inter-class feature entanglement is evident—particularly in regions highlighted by red rectangles—indicating poor discriminability. 
In contrast, Figure~\ref{lsq}(b) demonstrates that $L_{\mathrm{sq}}$ significantly increases inter-class margins and improves cluster separation. This visualization confirms that $L_{\mathrm{sq}}$ effectively guides the model toward learning more discriminative and semantically coherent representations.
\begin{table}[!t]
  \caption{Ablation Study on the Impact of the Coefficient $\alpha_1$ in the Loss Function $L = L_{vq} + \alpha_1 L_{sq}$}
  \centering
  % \small
  % \setlength{\tabcolsep}{10pt}
  \begin{tabular}{ccccccccc}
    \toprule
     $L_{sq}$ & Cur & M-O & Art & Con & Salt & M-W & Gen & Avg \\
    \midrule
    -   & 86.4 & 86.0 & 74.0 & 80.2 & 83.1 & 79.5 & 84.3 & 81.93 \\
    0.2 & 87.4 & 87.7 & 75.3 & 81.8 & 83.8 & 79.7 & 84.7 & 82.91 \\
    0.3 & 87.2 & 87.7 & 76.7 & 81.6 & 83.6 & 82.9 & 84.8 & \textbf{83.50} \\
    0.4 & 87.2 & 87.7 & 75.6 & 80.9 & 84.4 & 81.8 & 84.8 & 83.20 \\
    \bottomrule
    \end{tabular}
    \label{tab:lq}
\end{table}

\subsection{Limitations}
While OTSNet demonstrates strong robustness on complex and visually ambiguous scene text benchmarks (e.g., Union14M, OST), it exhibits a slight performance gap compared to language-model-heavy approaches (e.g., ABINet, MATRN) on clean, short-text datasets such as IIIT5K and IC13. 
This limitation stems from OTSNet’s design philosophy: it deliberately minimizes reliance on linguistic priors to avoid generating plausible but visually inconsistent predictions in open-world scenarios. 
Consequently, in ideal conditions where text is high-resolution, front-facing, and lexically common, the model forgoes the “guessing” advantage offered by strong language models, leading to marginally lower accuracy (e.g., 95.82\% vs. 96.29\% on Common Benchmarks). 
This trade-off reflects a conscious prioritization of visual fidelity over linguistic shortcut learning—a choice aligned with real-world deployment needs but suboptimal for synthetic or curated benchmarks. 
Future work may explore adaptive fusion mechanisms that dynamically adjust the visual–linguistic balance based on input complexity.

\section{Conclusions}\label{sec5}

In this paper, we propose the cognitively-inspired OTSNet to address the disjoint training problem in vision-language cross-modal collaboration.
Drawing on the human cognitive mechanism of observation-thinking-spelling, our framework coordinates multi-modal features through three progressive stages:
The observation stage employs the DAME to suppress irrelevant region interference through differential attention maps, thereby enhancing focal region localization accuracy.
Building upon this, the thinking stage dynamically reinforces semantic focus through PAM and SQ to obtain abstract glyph semantic representations.
Finally, the spelling stage introduces MMCV that constructs triple-constraint relationships among visual, semantic, and character features through a trinity verification network, endowing the model with self-rectifying prediction capabilities. 
Experimental results demonstrate that OTSNet significantly improves robustness in visually complex and occluded scenes, establishing a new state-of-the-art for cognition-inspired STR.

\section*{CRediT authorship contribution statement}
\textbf{Lixu Sun}: Conceptualization, Formal analysis, Methodology, Software, Validation, Investigation, Writing.
\textbf{Nurmemet Yolwas}: Resources, Supervision, Writing.
\textbf{Wushouer Silamu}: Resources, Supervision, Writing.

\section*{Declaration of competing interest}
The authors declare that they have no known competing financial interests or personal relationships that could have appeared to influence the work reported in this paper.

\section*{Acknowledgments}
This work was supported by the National Key Research and Development Program of China under Grant No. 2023B01005,
 titled \textit{Research and Application of Key Technologies for Intelligent Perception, Analysis, and Decision-Making in the Multimodal Internet Content Security Ecosystem}.

\section*{Data statement}
The ICDAR 2013 (IC13) \cite{karatzas2013icdar}, Street View Text (SVT) \cite{7}, IIIT5K Word (IIIT5K) \cite{IIIT5K}, ICDAR 2015 (IC15) \cite{karatzas2015icdar}, Street View Text Perspective (SVTP) \cite{phan2013recognizing}, CUTE80 \cite{risnumawan2014robust}, and the Union14M-L dataset \cite{jiang2023revisiting} are publicly available.


\bibliographystyle{elsarticle-num} 
\bibliography{ref}

\begin{thebibliography}{10}
\expandafter\ifx\csname url\endcsname\relax
  \def\url#1{\texttt{#1}}\fi
\expandafter\ifx\csname urlprefix\endcsname\relax\def\urlprefix{URL }\fi
\expandafter\ifx\csname href\endcsname\relax
  \def\href#1#2{#2} \def\path#1{#1}\fi

\bibitem{1}
Z.~Fu, H.~Xie, S.~Fang, Y.~Wang, M.~Xing, Y.~Zhang, Learning pixel affinity pyramid for arbitrary-shaped text detection, ACM Transactions on Multimedia Computing, Communications and Applications 19~(1s) (2023) 1--24.

\bibitem{3}
Z.~Liu, W.~Zhou, H.~Li, Mfecn: Multi-level feature enhanced cumulative network for scene text detection, ACM Trans. Multimedia Comput. Commun. Appl. 17~(3) (Jul. 2021).

\bibitem{6}
M.~Jaderberg, K.~Simonyan, A.~Vedaldi, A.~Zisserman, Reading text in the wild with convolutional neural networks, International journal of computer vision 116 (2016) 1--20.

\bibitem{8}
Y.~Wang, H.~Xie, Z.-J. Zha, M.~Xing, Z.~Fu, Y.~Zhang, Contournet: Taking a further step toward accurate arbitrary-shaped scene text detection, in: proceedings of the IEEE/CVF conference on computer vision and pattern recognition, 2020, pp. 11753--11762.

\bibitem{9}
S.~Fang, H.~Xie, Y.~Wang, Z.~Mao, Y.~Zhang, Read like humans: Autonomous, bidirectional and iterative language modeling for scene text recognition, in: Proceedings of the IEEE/CVF conference on computer vision and pattern recognition, 2021, pp. 7098--7107.

\bibitem{VisionLAN}
Y.~Wang, H.~Xie, S.~Fang, J.~Wang, S.~Zhu, Y.~Zhang, From two to one: A new scene text recognizer with visual language modeling network, in: Proceedings of the IEEE/CVF International Conference on Computer Vision, 2021, pp. 14194--14203.

\bibitem{wei2024busnet}
J.~Wei, H.~Zhan, Y.~Lu, X.~Tu, B.~Yin, C.~Liu, U.~Pal, Image as a language: Revisiting scene text recognition via balanced, unified and synchronized vision-language reasoning network, in: Proceedings of the AAAI Conference on Artificial Intelligence, Vol.~38, 2024, pp. 5885--5893.

\bibitem{ungerleider2000mechanisms}
S.~K. Ungerleider, L.~G, Mechanisms of visual attention in the human cortex, Annual review of neuroscience 23~(1) (2000) 315--341.

\bibitem{li2024volter}
J.-N. Li, X.-Q. Liu, X.~Luo, X.-S. Xu, Volter: Visual collaboration and dual-stream fusion for scene text recognition, IEEE Transactions on Multimedia 26 (2024) 6437--6448.

\bibitem{vits}
A.~Dosovitskiy, L.~Beyer, A.~Kolesnikov, D.~Weissenborn, X.~Zhai, T.~Unterthiner, M.~Dehghani, M.~Minderer, G.~Heigold, S.~Gelly, J.~Uszkoreit, N.~Houlsby, \href{https://openreview.net/forum?id=YicbFdNTTy}{An image is worth 16x16 words: Transformers for image recognition at scale}, in: International Conference on Learning Representations, 2021.
\newline\urlprefix\url{https://openreview.net/forum?id=YicbFdNTTy}

\bibitem{zhong2024ndorder}
D.~Zhong, H.~Zhan, S.~Lyu, C.~Liu, B.~Yin, P.~Shivakumara, U.~Pal, Y.~Lu, Ndorder: exploring a novel decoding order for scene text recognition, Expert Systems with Applications 249 (2024) 123771.

\bibitem{zheng2024cdistnet}
T.~Zheng, Z.~Chen, S.~Fang, H.~Xie, Y.-G. Jiang, Cdistnet: Perceiving multi-domain character distance for robust text recognition, International Journal of Computer Vision 132~(2) (2024) 300--318.

\bibitem{yang2024class}
M.~Yang, B.~Yang, M.~Liao, Y.~Zhu, X.~Bai, Class-aware mask-guided feature refinement for scene text recognition, Pattern Recognition 149 (2024) 110244.

\bibitem{laplante2018comprehensive}
P.~A. Laplante, R.~Cravey, L.~P. Dunleavy, J.~L. Antonakos, R.~LeRoy, J.~East, N.~E. Buris, C.~J. Conant, L.~Fryda, R.~W. Boyd, et~al., Comprehensive dictionary of electrical engineering, CRC Press, 2018.

\bibitem{ye2024differential}
T.~Ye, L.~Dong, Y.~Xia, Y.~Sun, Y.~Zhu, G.~Huang, F.~Wei, Differential transformer, arXiv preprint arXiv:2410.05258 (2024).

\bibitem{zhao2024decoder}
S.~Zhao, Y.~Du, Z.~Chen, Y.-G. Jiang, Decoder pre-training with only text for scene text recognition, in: Proceedings of the 32nd ACM International Conference on Multimedia, 2024, pp. 5191--5200.

\bibitem{zhang2016multi}
Z.~Zhang, C.~Zhang, W.~Shen, C.~Yao, W.~Liu, X.~Bai, Multi-oriented text detection with fully convolutional networks, in: Proceedings of the IEEE conference on computer vision and pattern recognition, 2016, pp. 4159--4167.

\bibitem{ctc}
A.~Graves, S.~Fern{\'a}ndez, F.~Gomez, J.~Schmidhuber, Connectionist temporal classification: labelling unsegmented sequence data with recurrent neural networks, in: Proceedings of the 23rd international conference on Machine learning, 2006, pp. 369--376.

\bibitem{shi2016end}
B.~Shi, X.~Bai, C.~Yao, An end-to-end trainable neural network for image-based sequence recognition and its application to scene text recognition, IEEE transactions on pattern analysis and machine intelligence 39~(11) (2016) 2298--2304.

\bibitem{bai2018edit}
F.~Bai, Z.~Cheng, Y.~Niu, S.~Pu, S.~Zhou, Edit probability for scene text recognition, in: proceedings of the IEEE conference on computer vision and pattern recognition, 2018, pp. 1508--1516.

\bibitem{bautista2022scene}
D.~Bautista, R.~Atienza, Scene text recognition with permuted autoregressive sequence models, in: European conference on computer vision, Springer, 2022, pp. 178--196.

\bibitem{shi2018aster}
B.~Shi, M.~Yang, X.~Wang, P.~Lyu, C.~Yao, X.~Bai, Aster: An attentional scene text recognizer with flexible rectification, IEEE transactions on pattern analysis and machine intelligence 41~(9) (2018) 2035--2048.

\bibitem{atienza2021vision}
R.~Atienza, Vision transformer for fast and efficient scene text recognition, in: International conference on document analysis and recognition, Springer, 2021, pp. 319--334.

\bibitem{qiao2021pimnet}
Z.~Qiao, Y.~Zhou, J.~Wei, W.~Wang, Y.~Zhang, N.~Jiang, H.~Wang, W.~Wang, Pimnet: a parallel, iterative and mimicking network for scene text recognition, in: Proceedings of the 29th ACM international conference on multimedia, 2021, pp. 2046--2055.

\bibitem{12}
D.~Yu, X.~Li, C.~Zhang, T.~Liu, J.~Han, J.~Liu, E.~Ding, Towards accurate scene text recognition with semantic reasoning networks, in: Proceedings of the IEEE/CVF conference on computer vision and pattern recognition, 2020, pp. 12113--12122.

\bibitem{du2022svtr}
Y.~Du, Z.~Chen, C.~Jia, X.~Yin, T.~Zheng, C.~Li, Y.~Du, Y.-G. Jiang, \href{https://doi.org/10.24963/ijcai.2022/124}{Svtr: Scene text recognition with a single visual model}, in: L.~D. Raedt (Ed.), Proceedings of the Thirty-First International Joint Conference on Artificial Intelligence, {IJCAI-22}, International Joint Conferences on Artificial Intelligence Organization, 2022, pp. 884--890, main Track.
\newblock \href {https://doi.org/10.24963/ijcai.2022/124} {\path{doi:10.24963/ijcai.2022/124}}.
\newline\urlprefix\url{https://doi.org/10.24963/ijcai.2022/124}

\bibitem{MATRN}
B.~Na, Y.~Kim, S.~Park, Multi-modal text recognition networks: Interactive enhancements between visual and semantic features, in: European Conference on Computer Vision, Springer, 2022, pp. 446--463.

\bibitem{luo2021separating}
C.~Luo, Q.~Lin, Y.~Liu, L.~Jin, C.~Shen, Separating content from style using adversarial learning for recognizing text in the wild, International Journal of Computer Vision 129 (2021) 960--976.

\bibitem{yang2016consistent}
W.~Yang, Y.~Tian, F.~Zhou, Q.~Liao, H.~Chen, C.~Zheng, Consistent coding scheme for single-image super-resolution via independent dictionaries, IEEE Transactions on Multimedia 18~(3) (2016) 313--325.

\bibitem{yang2018drfn}
X.~Yang, H.~Mei, J.~Zhang, K.~Xu, B.~Yin, Q.~Zhang, X.~Wei, Drfn: Deep recurrent fusion network for single-image super-resolution with large factors, IEEE Transactions on Multimedia 21~(2) (2018) 328--337.

\bibitem{wang2019textsr}
W.~Wang, E.~Xie, P.~Sun, W.~Wang, L.~Tian, C.~Shen, P.~Luo, Textsr: Content-aware text super-resolution guided by recognition, arXiv preprint arXiv:1909.07113 (2019).

\bibitem{li2021character}
M.~Li, B.~Fu, Z.~Zhang, Y.~Qiao, Character-aware sampling and rectification for scene text recognition, IEEE Transactions on Multimedia 25 (2021) 649--661.

\bibitem{wu2022two}
L.~Wu, Y.~Xu, J.~Hou, C.~P. Chen, C.-L. Liu, A two-level rectification attention network for scene text recognition, IEEE Transactions on Multimedia 25 (2022) 2404--2414.

\bibitem{lee2020recognizing}
J.~Lee, S.~Park, J.~Baek, S.~J. Oh, S.~Kim, H.~Lee, On recognizing texts of arbitrary shapes with 2d self-attention, in: Proceedings of the IEEE/CVF Conference on Computer Vision and Pattern Recognition Workshops, 2020, pp. 546--547.

\bibitem{cheng2018aon}
Z.~Cheng, Y.~Xu, F.~Bai, Y.~Niu, S.~Pu, S.~Zhou, Aon: Towards arbitrarily-oriented text recognition, in: Proceedings of the IEEE conference on computer vision and pattern recognition, 2018, pp. 5571--5579.

\bibitem{Conformer}
A.~Gulati, J.~Qin, C.-C. Chiu, N.~Parmar, Y.~Zhang, J.~Yu, W.~Han, S.~Wang, Z.~Zhang, Y.~Wu, R.~Pang, Conformer: Convolution-augmented transformer for speech recognition, in: Interspeech 2020, 2020, pp. 5036--5040.
\newblock \href {https://doi.org/10.21437/Interspeech.2020-3015} {\path{doi:10.21437/Interspeech.2020-3015}}.

\bibitem{vaswani2017attention}
A.~Vaswani, N.~Shazeer, N.~Parmar, J.~Uszkoreit, L.~Jones, A.~N. Gomez, {\L}.~Kaiser, I.~Polosukhin, Attention is all you need, Advances in neural information processing systems 30 (2017).

\bibitem{jang2016categorical}
E.~Jang, S.~Gu, B.~Poole, \href{https://openreview.net/forum?id=rkE3y85ee}{Categorical reparameterization with gumbel-softmax}, in: International Conference on Learning Representations, 2017.
\newline\urlprefix\url{https://openreview.net/forum?id=rkE3y85ee}

\bibitem{karatzas2013icdar}
D.~Karatzas, F.~Shafait, S.~Uchida, M.~Iwamura, L.~G. i~Bigorda, S.~R. Mestre, J.~Mas, D.~F. Mota, J.~A. Almazan, L.~P. De~Las~Heras, Icdar 2013 robust reading competition, in: 2013 12th international conference on document analysis and recognition, IEEE, 2013, pp. 1484--1493.

\bibitem{7}
K.~Wang, B.~Babenko, S.~Belongie, End-to-end scene text recognition, in: 2011 International conference on computer vision, IEEE, 2011, pp. 1457--1464.

\bibitem{IIIT5K}
A.~Mishra, K.~Alahari, C.~Jawahar, Scene text recognition using higher order language priors, in: BMVC-British machine vision conference, BMVA, 2012.

\bibitem{karatzas2015icdar}
D.~Karatzas, L.~Gomez-Bigorda, A.~Nicolaou, S.~Ghosh, A.~Bagdanov, M.~Iwamura, J.~Matas, L.~Neumann, V.~R. Chandrasekhar, S.~Lu, et~al., Icdar 2015 competition on robust reading, in: 2015 13th international conference on document analysis and recognition (ICDAR), IEEE, 2015, pp. 1156--1160.

\bibitem{phan2013recognizing}
T.~Q. Phan, P.~Shivakumara, S.~Tian, C.~L. Tan, Recognizing text with perspective distortion in natural scenes, in: Proceedings of the IEEE international conference on computer vision, 2013, pp. 569--576.

\bibitem{risnumawan2014robust}
A.~Risnumawan, P.~Shivakumara, C.~S. Chan, C.~L. Tan, A robust arbitrary text detection system for natural scene images, Expert Systems with Applications 41~(18) (2014) 8027--8048.

\bibitem{jiang2023revisiting}
Q.~Jiang, J.~Wang, D.~Peng, C.~Liu, L.~Jin, Revisiting scene text recognition: A data perspective, in: Proceedings of the IEEE/CVF international conference on computer vision, 2023, pp. 20543--20554.

\bibitem{sheng2019nrtr}
F.~Sheng, Z.~Chen, B.~Xu, Nrtr: A no-recurrence sequence-to-sequence model for scene text recognition, in: 2019 International conference on document analysis and recognition (ICDAR), IEEE, 2019, pp. 781--786.

\bibitem{wang2020decoupled}
T.~Wang, Y.~Zhu, L.~Jin, C.~Luo, X.~Chen, Y.~Wu, Q.~Wang, M.~Cai, Decoupled attention network for text recognition, in: Proceedings of the AAAI conference on artificial intelligence, Vol.~34, 2020, pp. 12216--12224.

\bibitem{yu2020towards}
D.~Yu, X.~Li, C.~Zhang, T.~Liu, J.~Han, J.~Liu, E.~Ding, Towards accurate scene text recognition with semantic reasoning networks, in: Proceedings of the IEEE/CVF conference on computer vision and pattern recognition, 2020, pp. 12113--12122.

\bibitem{11}
Z.~Qiao, Y.~Zhou, D.~Yang, Y.~Zhou, W.~Wang, Seed: Semantics enhanced encoder-decoder framework for scene text recognition, in: Proceedings of the IEEE/CVF conference on computer vision and pattern recognition, 2020, pp. 13528--13537.

\bibitem{yue2020robustscanner}
X.~Yue, Z.~Kuang, C.~Lin, H.~Sun, W.~Zhang, Robustscanner: Dynamically enhancing positional clues for robust text recognition, in: European conference on computer vision, Springer, 2020, pp. 135--151.

\bibitem{MGP-STR}
P.~Wang, C.~Da, C.~Yao, Multi-granularity prediction for scene text recognition, in: European Conference on Computer Vision, Springer, 2022, pp. 339--355.

\bibitem{LPV-B}
B.~Zhang, H.~Xie, Y.~Wang, J.~Xu, Y.~Zhang, Linguistic more: Taking a further step toward efficient and accurate scene text recognition, arXiv preprint arXiv:2305.05140 (2023).

\bibitem{LISTER}
B.~Zhang, H.~Xie, Y.~Wang, J.~Xu, Y.~Zhang, \href{https://doi.org/10.24963/ijcai.2023/189}{Linguistic more: Taking a further step toward efficient and accurate scene text recognition}, in: E.~Elkind (Ed.), Proceedings of the Thirty-Second International Joint Conference on Artificial Intelligence, {IJCAI-23}, International Joint Conferences on Artificial Intelligence Organization, 2023, pp. 1704--1712, main Track.
\newblock \href {https://doi.org/10.24963/ijcai.2023/189} {\path{doi:10.24963/ijcai.2023/189}}.
\newline\urlprefix\url{https://doi.org/10.24963/ijcai.2023/189}

\bibitem{OTE}
J.~Xu, Y.~Wang, H.~Xie, Y.~Zhang, Ote: exploring accurate scene text recognition using one token, in: Proceedings of the IEEE/CVF Conference on Computer Vision and Pattern Recognition, 2024, pp. 28327--28336.

\bibitem{du2024svtrv2}
Y.~Du, Z.~Chen, H.~Xie, C.~Jia, Y.-G. Jiang, Svtrv2: Ctc beats encoder-decoder models in scene text recognition, arXiv preprint arXiv:2411.15858 (2024).

\end{thebibliography}




% %% For numbered reference style

%   \textit{\LaTeX: a document preparation system},
%   Addison Wesley, Massachusetts,

% \end{thebibliography}
\end{document}